\pdfoutput=1

\documentclass[11pt]{article}

\usepackage{emnlp2021}

\usepackage{times}
\usepackage{latexsym}
\usepackage{graphics}
\usepackage{graphicx}
\usepackage[figuresright]{rotating}
\usepackage{multirow}
\usepackage{makecell}
\usepackage{lscape}
\usepackage{setspace}
\usepackage{amssymb}
\usepackage[T1]{fontenc}

\usepackage[utf8]{inputenc}

\usepackage{microtype}

\title{Coreference Resolution for the Biomedical Domain: A Survey}

\author{Pengcheng Lu \and
  Massimo Poesio\\

School of Electronical Engineering and
Computer Science\\Queen Mary University of London, United Kingdom\\
  \texttt{\{pengcheng.lu, m.poesio\}@qmul.ac.uk
  }
}
\begin{document}
\maketitle
\begin{abstract}
Issues with coreference resolution are one of the most frequently mentioned  challenges for information extraction from the biomedical literature.
Thus, the biomedical genre has long been the second most researched genre for coreference resolution after the news domain, and the subject of a great deal of research for NLP in general.
In recent years this interest has grown enormously leading to the development of a number of substantial datasets, of domain-specific contextual language models, and of several architectures. 
In this paper we review the state-of-the-art of coreference in the biomedical domain with a particular attention on these most recent developments. 
\end{abstract}

\section{Introduction}

Coreference resolution is the process of identifying entities in a text and finding all mentions that refer to the same entities. 
It is a fundamental and challenging NLP task,  supporting downstream tasks such as information extraction and question answering.

In the biomedical domain, issues with coreference resolution are one of the most frequently mentioned challenges for information extraction from the biomedical literature (\citealt{castano2002anaphora}, \citealt{miwa2012boosting}).
Biomedical coreference resolution has become an essential task to support the discovery of complex information by identifying  coreference links in biomedical texts.

In recent years in particular, biomedical coreference resolution has attracted a great deal of attention due both to its great potential for application, and to its theoretical interest e.g., as an application of knowledge embeddings and entity linking. 
Several biomedical coreference corpora have been made available, especially for protein coreference which is a supporting task for BioNLP 2011 shared task \citep{nguyen2011overview}.

\begin{figure}[h]
\centering
\includegraphics[width=8cm,height=7.5cm]{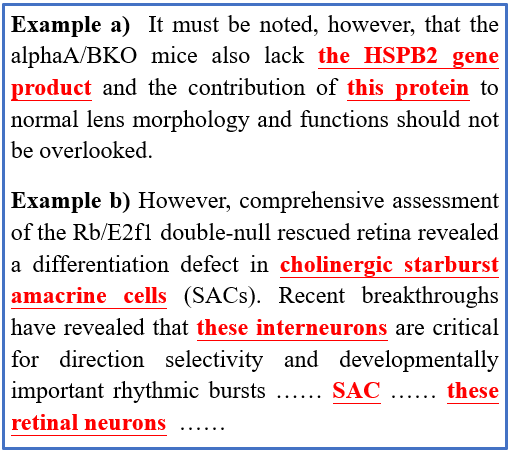}
\caption{Examples of coreference relations in the CRAFT-CR dataset. The mentions marked red are coreferent.}
\label{examples}
\end{figure}

Biomedical coreference is quite different from the general domain coreference, such as different markable types. Therefore, domain-specific knowledge is important for bridging the gap. Figure \ref{examples} shows two examples of biomedical coreference in the CRAFT-CR dataset \citep{cohen2017coreference}. In example a), the mention \emph{this protein} refers to the antecedent \emph{the HSPB2 gene product}. To understand the coreference relation, we need the background knowledge that proteins are fundamental encoded by genes. In example b), to correctly resolve the coreference relation between these different biomedical entities, the biomedical-domain knowledge that \emph{cholinergic starburst amacrine cell} is a kind of \emph{interneuron} and belongs to \emph{retinal neuron} are required.

A large number of coreference models for the biomedical domain have already been proposed, from rule-based models (\citealt{castano2002anaphora}, \citealt{kim2004bioar}, \citealt{lin2004pronominal}, \citealt{nguyen2012improving}, \citealt{miwa2012boosting}, \citealt{kilicoglu2016bio}, \citealt{li2018set}), machine learning-based models (\citealt{yang2004np}, \citealt{torii2005anaphora}, \citealt{su2008coreference}, \citealt{gasperin2009statistical}, \citealt{kim2011taming}) to recent deep learning-based models (\citealt{trieu2018investigating}, \citealt{trieu2019coreference}, \citealt{li2021knowledge}).  These models usually integrate biomedical specific information, 
typically specific rules, pre-trained embeddings and features.

This paper reviews and analyses coreference datasets and models for the biomedical domain, as well as recent biomedical language representation models which can enhance coreference models with domain-specific knowledge. In addition, we conduct experiments to evaluate the ability of these language represetation models for biomedical coreference task.

The structure of this paper is  as follows. 
In Section \ref{background} we briefly provide some background on coreference resolution in the general domain. 
Section \ref{datasets} reviews the main datasets used to study  biomedical coreference. 
Overviews of biomedical language representation models and biomedical coreference models are provided in Sections \ref{sec:bioLM} and  \ref{sec:coref_model}, respectively. 
Section \ref{sec:method} introduces the methodology of comparing the biomedical language representation models for coreference. Section \ref{results} presents the evaluation results including the performance of previous models and our experiments, and Section \ref{conclusion} concludes.

\section{Background}
\label{background}
Coreference resolution in the general domain has a long history of being studied from early heuristic-based and rule-based approaches to recent learning-based approaches.

\citet{lee2017end} proposed the first end-to-end neural coreference resolution model which uses LSTM encoder. Based on the end-to-end model, many extensions to the model have been proposed. BERT and SpanBERT were proposed to replace the LSTM encoder and achieved better performance on OntoNotes dataset (\citealt{joshi2019bert}, \citealt{joshi2020spanbert}). \citet{wu2020corefqa} adapted question-answering framework
on coreference resolution, and achieved the state-of-the-art result with 83.1\% F1 score on OntoNotes dataset. \citet{ye2020coreferential} proposed a novel language representation model CorefBERT, which can capture the coreferential relations in context.

However, these general coreference systems do not work well in the biomedical domain due to the lack of domain knowledge. For example, the end-to-end model \citep{lee2017end} only achieved 33.85\% and 61.25\% F1 scores on CRAFT-CR and BioNLP datasets respectively \citep{trieu2018investigating}, but achieved 68.8\% F1 score on OntoNotes dataset \citep{hovy2006ontonotes}, which covers multiple genres, such as newswire, broadcast news and web data.

\section{Biomedical Coreference Datasets}
\label{datasets}
Several biomedical datasets with coreference annotations exist, but different document selection criteria, annotation schemes, domains and coreference types were used. 
The best known include:

\emph{MEDSTRACT} \citep{pustejovsky2002medstract} is a corpus consisting of MEDLINE abstracts with coreference annotation. It is mainly concerned with two forms of anaphora:  pronominal and sortal (definite noun phrase) anaphora. 
This corpus adapted the MUC-7 annotation scheme \citep{hirschman1997muc};
in addition, 
semantic types from UMLS \citep{bodenreider2004unified} were also annotated. 

\emph{FlySlip} \citep{gasperin2007annotation} contains anaphoric links among  noun phrases, including coreferent and associative relations. 
Different from MEDSTRACT, full-text biomedical articles were annotated in this corpus. 
FlySlip was annotated according to a domain-specific annotation scheme.

\emph{GENIA-MedCo} \citep{su2008coreference} is a coreferentially annotated version of the GENIA corpus \citep{kim2003genia}, which in turn 
consists 
of 1999 MEDLINE abstracts. This corpus follows the MUC-7 annotation scheme, but adds more linguistic based relations.

\emph{DrugNerAR} \citep{segura2010resolving} was created to
study 
anaphoric expressions in the task of extracting drug-drug interactions in pharmacological literature. This corpus consists of 49 full-text from the DrugBank database, which contains 
4900 drug entries. 

\emph{BioNLP-ST’11 COREF} \citep{nguyen2011overview} 
was created in support of one of the tasks 
of 
the
BioNLP 2011 shared task, focusing on finding anaphoric protein references, and based on the observation that one of major difficulties in event extraction is coreference resolution. This corpus was derived from three resources: MedCo coreference annotation \citep{su2008coreference}, Genia event annotation \citep{kim2008corpus}, and Genia Treebank \citep{tateisi2005syntax}. 

\emph{HANAPIN} \citep{batista2011building} is comprised of 20 full-text articles from biochemistry literature. In addition to nominal and

\begin{landscape}
\renewcommand\arraystretch{1.15} 
\begin{table}
\centering
\resizebox{\linewidth}{!}{

\begin{tabular}{lllllllc}
\hline
\textbf{Dataset} & \textbf{Document Type} & \textbf{Annotation} & \textbf{Domain} &  \textbf{Coreference type} & \textbf{Semantic type of} &  \textbf{Relation Type} & \textbf{Publicly}\\
 &  & \textbf{Scheme} &  &  & \textbf{markables} & & \textbf{Available}\\
\hline
MEDSTRACT & 100 MEDLINE abstracts & MUC-7 & molecular biology & pronoun, & UMLS semantic types & pairwise coreference & Yes\\
 & & & & sortal & & & \\
FlySlip & 5 full-text articles & domain-specific & fruit fly genomics & noun phrase & genetic entities & coreference chain & Yes\\
GENIA-MedCo & 1999 MEDLINE abstracts & MUC-7 & transcription factors & pronoun, & GENIA Ontology types & pairwise coreference & Yes\\
 & & & in human blood cells & noun phrase & & & \\
DrugNerAR & 49 full articles & MUC-7 & drug-drug interactions & pronoun, & drugs & pairwise coreference & Yes\\
 & & & &  noun phrase & & & \\
BioNLP 11 COREF & 1210 MEDLINE abstracts & BioNLP & molecular biology & pronoun, & protein names & pairwise coreference & Yes\\
 & & & & noun phrase & & & \\
HANAPIN & 20 full-text articles & MedCo & marine natural & pronoun, & chemical compounds, & coreference chain & Yes\\
 & & & products chemistry & sortal, & organisms, & & \\
 & & & & numerical, & drug effects, & & \\
 & & & & abbreviation & diseases, & & \\
 & & & & & drug targets & & \\
CRAFT-CR & 97 full texts articles & OntoNotes &  mouse genomics & pronoun, & all types & coreference chain & Yes\\
 & & & & noun phrase, & & & \\
 & & & & verb, & & & \\
 & & & & event, & & & \\
 & & & & Nominal premodifiers & & & \\
\hline
\end{tabular}

}
\caption{\label{tab:datasets}
Comparison of biomedical datasets with coreferent annotations.}
\end{table}
\end{landscape}

\noindent pronominal anaphora, this corpus also annotated abbreviation/acronyms and numerical anaphora.

\emph{CRAFT-CR} \citep{cohen2017coreference} consists of 97 full-text biomedical journal articles. Similar to the general domain, this corpus was annotated with coreferent chains in full-text articles, while most other biomedical coreference datasets focuse on annotating the pairwise coreference relation between an anaphor and its antecedent. In addition, all coreference expressions were annotated regardless of semantic type.

These datasets are summarized in Table \ref{tab:datasets}.

\section{Biomedical Language Representation Models}
\label{sec:bioLM}

The news domain and the biomedical domain 
are  different in a number of respects,  
such as markable types. 
Some authors have argued that biomedical domain knowledge is the key to bridging the gap \citep{choi2014evaluation},
and that therefore, incorporating biomedical specific representation is beneficial for resolving coreferring expressions in the biomedical domain. 
In this section, we will give a brief introduction to biomedical language representation models. 


\subsection{Pre-training on biomedical corpora}
Following the success of large-scale pre-training language models (PLMs) in the general domain, several biomedical-domain PLMs have been developed in recent years by pre-training on large-scale biomedical corpora.

Most 
biomedical PLMs conduct continual pretraining of the general domain PLMs and still use vocabulary trained on the general domain text. BioBERT \citep{lee2020biobert} is the first transformer-based biomedical PLM, pre-trained on PubMed abstracts and PubMed Central full-text articles. 
ClinicalBERT and Bio\_ClinicalBERT \citep{alsentzer2019publicly} are pre-trained on MIMIC-III Clinical Notes, whereas BlueBERT \citep{peng2019transfer} uses both PubMed and MIMIC-III for pre-training. All these models are pre-trained based on general BERT, except Bio\_ClinicalBERT which is initialized from BioBERT.

 In addition to initializing from general BERT, some biomedical PLMs are directly pre-trained on biomedical text from scratch and use domain-specific custom vocabulary.
 SciBERT \citep{beltagy2019scibert} is pre-trained on biomedical and computer science papers from scratch and achieved good performance on many scientific NLP tasks. PubMedBERT \citep{gu2020domain} and BioELECTRA \citep{raj2021bioelectra} are both pre-trained on PubMed abstract and PubMed Central full text articles, but the latter adopts ELECTRA architecture \citep{clark2019electra}. 
 BioMegatron \citep{shin2020bio} is a large-scale model based on Megatron \citep{shoeybi2019megatron} architecture. It also investigated the effect of vocabulary and corpora domain on the performance of biomedical tasks.
 
\subsection{Integrating biomedical knowledge bases}
Although the biomedical PLMs, such as BioBERT, have achieved good performance on many biomedical tasks, however, these models can be further enhanced by integrating biomedical knowledge bases, such as UMLS \citep{bodenreider2004unified}.

Several models enhance biomedical PLMs by integrating synonym knowledge from UMLS. Each mention in the biomedical text can be linked to a Concept Unique Identifier (CUI) in UMLS, and each CUI has a synonym set. SAPBERT \citep{liu2021self}, UMLSBERT \citep{michalopoulos2021umlsbert} and BIOSYN \citep{sung2020biomedical} further pre-trained PubMedBERT, Bio\_Clinical BERT and BioBERT on UMLS synonyms, using multi-similarity loss, multi-label loss and synonym
marginalization algorithm respectively.

In addition to synonym knowledge, Clinical KB-BERT \citep{hao2020enhancing} injects UMLS relation knowledge into BioBERT. Whereas CODER \citep{yuan2020coder} learns both synonym and relation knowledge based on PubMedBERT or mBERT \citep{devlin2019bert} via contrastive learning. Also, some research focus on fusing the UMLS entity embeddings with contextual embeddings to improve biomedical PLMs (\citealp{he2020integrating}; \citealp{fei2021enriching}; \citealp{yuan2021improving}).

This paper selected some of the models above to evaluate the ability of biomedical-specific representation for biomedical coreference task, detailed in Section \ref{sec:method}.

\section{Coreference Models for the Biomedical Domain}
\label{sec:coref_model}

\subsection{Rule-based models}
Early approaches to biomedical coreference resolution are primarily rule-based. These models rely on syntactic parsers to extract hand-crafted features and rules. 

\citet{nguyen2012improving} implemented a protein coreference system that makes use of syntactic information from the parser output, and protein-indicated information. The results showed that domain-specific semantic information is important for coreference resolution. \citet{miwa2012boosting} developed a rule-based coreference system, as a part of the EventMine event extraction system. A set of rules was developed based on syntactic trees and predicate-argument structures. The system achieved 55.9\% F1 score on BioNLP 2011 protein coreference task. \citet{kilicoglu2016bio} developed a new corpus of structured drug labels and proposed a general framework based on a smorgasbord architecture for fine-grained biomedical coreference resolution. The framework adopted different strategies for each coreference type and mention type, and combined them to reach desired performance, like selecting dishes from a smorgasbord. \citet{li2018set} presented two methods for bio-entity coreference resolution: a rule-based method and a recurrent neural network (RNN) model. The rule-based model created a set of syntactic rules or semantic constraints for
coreference and achieved a state-of-the-art performance with 62.0\% F1 score on BioNLP 2011 protein coreference task. 

These rule-based models mostly designed rules for specific type of coreference relation and even specific corpus, which limits the scope of the resolution.

\subsection{Machine learning-based models}
In the early years, due to the lack of publicly available annotated corpora, researchers have to annotate their own corpora for developing machine learning approaches (\citealp{yang2004np}, \citealp{torii2005anaphora}, \citealp{su2008coreference}, \citealp{gasperin2009statistical}).

After the BioNLP 2011 protein coreference dataset was made publicly available, several machine learning-based models were developed for this task. \citet{kim2011taming} adapted a general coreference system Reconcile \citep{stoyanov2010coreference} for the biomedical domain by modifying several components to biomedical texts. It trained two separate classifiers for detecting anaphora and
antecedent mentions.

In addition to using machine learning-based methods only, several models adopted hydrid approach, i.e., combining both machine learning-based and rule-based methods. \citet{d2012anaphora} proposed a hybrid approach that used a classifier with syntactic path-based features. It investigated five different learning-based methods, and a rule-based approach for anaphora resolution. This model achieved a superior performance than previous either rule-based or learning-based models on BioNLP 2011 protein coreference task.
\citet{li2014coreference} later also used a hybrid approach, adopting the rule-based method or the machine learning method for three types of anaphora. As the method of \citet{d2012anaphora}, they also used different rules for different types of anaphora. The system achieved better performance with 68.6\% F1 score than previous methods on BioNLP 2011 protein coreference development data.

\begin{table}
\centering
\begin{tabular}{lll}
\hline
\textbf{} & \textbf{BioNLP} & \textbf{CRAFT}\\
\hline
Training set (docs) & 800 & 60\\
Development set (docs) & 150 & 7\\
Test set (docs) & 260 & 30\\
\hline
Avg. sent. per doc & 9.2 & 312.4\\
Avg. words per doc & 258.0 & 8181.0\\
\hline
\end{tabular}

\caption{\label{statistics}
Statistics of BioNLP and CRAFT.
}
\end{table}

\begin{table*}
\centering
\resizebox{\linewidth}{!}{
\begin{tabular}{ccclllllll}
\hline
\textbf{} & \textbf{Classification} & \textbf{} & \textbf{Model} & \textbf{} & \textbf{dev} & \textbf{} & \textbf{} & \textbf{test} & \textbf{}\\
\textbf{rule-} & \textbf{machine} & \textbf{deep} & \textbf{} & \textbf{R}  & \textbf{P}  & \textbf{F1}  & \textbf{R}  & \textbf{P}  & \textbf{F1}\\
\textbf{based} & \textbf{learning} & \textbf{learning} & \textbf{} &  &   &   &  &  & \\
\hline
 & $\checkmark$ & & Reconcile \citep{kim2011taming} & 26.7 & 74.0 & 39.3 & 22.2 & 73.3 & 34.1\\
$\checkmark$ & & & \citep{nguyen2012improving} & 57.8 & 67.8 & 62.4 & 52.5 & 50.2 & 51.3\\
$\checkmark$ & & & EventMine \citep{miwa2012boosting} & 53.5 & 69.8 & 60.5 & 50.4 & 62.7 & 55.9\\
$\checkmark$ & $\checkmark$ & & \citep{d2012anaphora} & 59.9 & 77.1 & 67.4 & 55.6 & 67.2 & 60.9\\
$\checkmark$ & $\checkmark$ & & \citep{li2014coreference} & 69.8 & 67.5 & 68.6 &  - &  - &  - \\
$\checkmark$ & & & Simple system \citep{choi2016categorical} & 64.4 & 63.4 & 63.9 & 50.0 & 46.3 & 48.1\\
$\checkmark$ & & & \citep{kilicoglu2016bio} & 63.2 & 72.4 & 67.5 & - & - & - \\
$\checkmark$ & & & \citep{li2018set}-rule & 68.8 & 76.0 & \textbf{72.2} & 60.2 & 63.8 & 62.0\\
 & & $\checkmark$ & \citep{li2018set}-neural & 60.4 & 61.9 & 61.2 & 54.9 & 58.0 & 56.4\\
 & & $\checkmark$ & E2E\_MetaMap \citep{trieu2018investigating} & 56.7 & 71.7 & 63.1 & 47.5 & 55.6 & 51.2\\
 & & $\checkmark$ & KB-attention \citep{li2021knowledge} & 63.4 & 68.1 & 65.6 & 69.4 & 69.6 & \textbf{69.5}\\
\hline
\end{tabular}
}
\caption{\label{BioNLP}
Performance of biomedical coreference models on BioNLP 2011 protein correference development and test sets.
}
\end{table*}

\subsection{Deep learning-based models}
In recent years, much effort has been made on using deep learning methods for biomedical coreference.

\citet{trieu2018investigating} applied general domain end-to-end neural coreference resolution
system \citep{lee2017end} to biomedical text, integrating the domain specific features to enhance the system. The model was evaluated on BioNLP 2011 protein coreference dataset and CRAFT-CR dataset. The results indicated that in-domain embeddings and domain-specific features helped improve the performance. Then, 
\citet{trieu2019coreference} proposed a system to address the challenge of coreference resolution in the full-text articles in the CRAFT-CR dataset. The model also applied end-to-end system \citep{lee2017end}, but enhanced the system by utilizing a syntax-based mention filtering method and replacing LSTM with BERT. This model achieved better performance on the CRAFT-CR dataset.

Different from the models above, \citet{li2021knowledge} integrated external knowledge to enhance the neural coreference system for biomedical texts. A knowledge attention module was developed to select the most related and helpful knowledge triplets. This model achieved the state-of-the-art performance on the BioNLP 2011 protein coreference dataset and CRAFT-CR dataset.

\section{Comparing the Biomedical Language Representation Models for Coreference}
\label{sec:method}

In Section \ref{sec:bioLM}, we introduced a series of biomedical language representation models. To investigate the ability of these models for biomedical coreference task, we conduct experiments to evaluate these models on CRAFT-CR dataset.

\subsection{Baseline model}
We employ the higher-order coreference model \citep{lee2018higher} as the baseline model, but use different pre-trained language models with BERT architecture to replace LSTM encoder.

The goal is to learn a distribution $P (y_{i}$) over possible antecedents $Y(i)$ for each span $i$:
\begin{equation}
P(y_{i})=\frac{e^{s(i,y_{i})}}{\sum_{y^{'}\in Y(i)}e^{s(i,y^{'})}}
\end{equation}
where $s(i, j)$ is a pairwise score for a coreference link between span $i$ and span $j$. The pairwise score is computed by the mention score of $i$, the mention score of
$j$, and two kinds of joint compatibility scores of $i$ and $j$:
\begin{equation}
s(i,j)=s_{m}(i)+s_{m}(j)+s_{c}(i,j)+s_{a}(i,j)
\end{equation}
The mention score and joint compatibility scores are computed using span representation $g_i$ and $g_j$ from bidirectional LSTMs: 
\begin{equation}
s_{m}(i)=\emph{FFNN}_{m}(g_{i})
\end{equation}
\begin{equation}
s_{c}(i,j)=g_{i}^{T}W_{c}\ g_{j}
\end{equation}
\begin{equation}
s_{a}(i,j)=\emph{FFNN}_{a}([g_{i},g_{j},g_{i}\circ g_{j},\phi (i,j)])
\end{equation}
where \emph{FFNN(·)} represents a feed-forward neural network, $W_c$ is a learned weight matrix, ◦ denotes element-wise product, and $\phi (i,j)$ represents speaker and metadata features.

\subsection{Applying pre-trained language models}
We apply two types of PLM to replace LSTM encoder respectively: 

\emph{Biomedical PLMs}: to enhance the baseline model with biomedical domain knowledge, several biomedical PLMs are selected, including models pre-training on biomedical corpora or integrating biomedical knowledge bases.

\emph{SpanBERT}: since SpanBERT \citep{joshi2020spanbert} is a state-of-the-art coreference resolution model in the general domain, we also evaluate SpanBERT and general BERT \citep{joshi2019bert} on biomedical coreference.

Since CRAFT-CR is a more challenging biomedical coreference dataset consisting of full-text articles, we choose CRAFT-CR to fine-tune and evaluate these models. The details are introduced in Section \ref{sec:eval}.

\begin{table*}
\centering
\begin{tabular}{llllllll}
\hline
\textbf{Model} & \textbf{$B^3$} & \textbf{BLANC} & \textbf{CEAFE} & \textbf{CEAFM} & \textbf{LEA} & \textbf{MUC} & \textbf{Avg.}\\
\hline
E2E\_MetaMap \citep{trieu2018investigating} & 36.4 & 46.5 & 33.1 & 41.0 & 32.4 & 51.8 & 40.2\\
BERT\_filter \citep{trieu2019coreference} & 44.0 & 48.9 & 39.8 & 49.0 & 40.0 & 57.0 & 46.4\\
KB-attention \citep{li2021knowledge} & 54.9 & 63.1 & 48.6 & 59.4 & 51.3 & 	64.5 & \textbf{57.0}\\
\hline
\end{tabular}

\caption{\label{CRAFT}
F1 scores of biomedical coreference models on CRAFT-CR test set.
}
\end{table*}

\begin{table*}
\centering
\resizebox{\linewidth}{!}{
\begin{tabular}{llllllll}
\hline
\textbf{Model} & \textbf{$B^3$} & \textbf{BLANC} & \textbf{CEAFE} & \textbf{CEAFM} & \textbf{LEA} & \textbf{MUC} & \textbf{Avg.}\\
\hline
BioBERT & 41.67 & 42.39 & 32.44 & 45.15 & 39.00
 & 53.66 & 42.38\\
SciBERT & 25.66 & 28.30 & 16.76 & 30.34 & 22.67 & 40.70 & 27.41 \\
Bio\_ClinicalBERT & 38.19 & 36.91 & 30.11 & 41.56 & 35.57
 & 48.22 & 38.43\\
PubMedBERT & 34.96 & 33.14 & 25.49 & 38.49 & 32.32
 & 47.02 & 35.24\\
UMLSBERT & 27.53 & 26.95 & 19.95 & 31.40 & 24.95
 & 39.80 & 28.43\\
Clinical KB-BERT & 44.56 & 44.99 & 37.25 & 48.29 & 41.67
 & 55.17 & 45.32\\
\hline
BERT\_base & 32.96 & 31.36 & 22.58 & 36.25 & 30.73
 & 44.34 & 33.04\\
SpanBERT\_base & 47.05 & 46.30 & 39.90 & 51.27 & 44.36
 & 57.69 & \textbf{47.76}\\
\hline
\end{tabular}
}
\caption{\label{experiment}
F1 scores of different PLMs combined with c2f-coref model on CRAFT-CR test set.
}
\end{table*}

\section{Results}
\label{results}
In this section, we first present the performance achieved by previous biomedical coreference models decribed in Section \ref{sec:coref_model}. Then we describe our experiment and report the results.

\subsection{Results by datasets}
Recent biomedical coreference models are mostly evaluated on BioNLP 2011 protein coreference dataset\footnote{\url{http://2011.bionlp-st.org/home/protein-gene-coreference-task}} and CRAFT-CR dataset\footnote{\url{https://github.com/UCDenver-ccp/craft-shared-tasks}}, of which the statistics are shown in Table \ref{statistics}.
The performance on the two datasets are summarized and analysed respectively as follows.

Table \ref{BioNLP} shows the performance of different biomedical coreference models on BioNLP 2011 protein coreference development and test sets. These models are evaluated using the scorer provided by the BioNLP shared task organisers. As shown in Table \ref{BioNLP}, KB-attention \citep{li2021knowledge} achieved the best performance of 69.5\% F1 score on the test set of BioNLP. This indicates that integrating external biomedical knowledge base can further enhance the coreference models for the biomedical domain. In addition, compared with deep learning-based models, some rule-based (\citealt{kilicoglu2016bio}, \citealt{li2018set})  or hybrid models (\citealt{d2012anaphora}, \citealt{li2014coreference}) still achieved favorable performance.

Table \ref{CRAFT} shows the F1 scores of different biomedical coreference models on CRAFT-CR test set. We can see that the best performance is also achieved by KB-attention \citep{li2021knowledge}, showing the advantage of fine-grained knowledge base integration. However, the results on CRAFT-CR are overall lower than those on BioNLP. The possible reason is that CRAFT-CR consists of full-text articles, hence the length of documents in CRAFT-CR is much greater. This makes CRAFT-CR more challenging than BioNLP dataset which comprises abstracts only.

\subsection{Experiments}
\label{sec:eval}
\subsubsection{Experimental setup}
We conduct experiment using following models:
\begin{itemize}
\item[-]\textbf{biomedical PLMs+c2f-coref}:
we refer to the higher-order coreference model \citep{lee2018higher} as \emph{c2f-coref}. We build the c2f-coref system on top of different biomedical PLMs respectively, including BioBERT, SciBERT, Bio\_ClinicalBERT, PubMedBERT, UMLSBERT, and Clinical KB-BERT. Among these models, UMLSBERT and Clinical KB-BERT integrate external biomedical knowledge base, i.e., UMLS, while other models are pre-trained on large-scale biomedical datasets. 

\item[-]\textbf{BERT\_base+c2f-coref} \citep{joshi2019bert}: the c2f-coref system on top of BERT representation.

\item[-]\textbf{SpanBERT\_base+c2f-coref} \citep{joshi2020spanbert}: the c2f-coref system on top of SpanBERT\_base, which pre-trained span representations to better represent and predict spans of text.
\end{itemize}

\begin{table*}
\centering
\resizebox{\linewidth}{!}{
\begin{tabular}{llllllll}
\hline
\textbf{Model} & \textbf{Evaluation script} & \textbf{Programming} & \textbf{Time cost} & \textbf{MUC} & \textbf{$B^3$} & \textbf{CEAFE} & \textbf{Avg.}\\
 &  & \textbf{language} &  & &  &  &\\
\hline
\multirow{3}{*}{\makecell[tl]{SpanBERT\_base\\+c2f-coref}}  & CoNLL scorer 9.0 & Perl & about 1.5h & 48.67 & 8.61
 & 18.90 & 25.39\\
 & CoVal script & Python & about 30s & 55.99 & 43.97 & 40.01 & 46.66\\
 & CRAFT evaluation script & Clojure & about 20m & 57.69 & 47.05
 & 39.90 & 48.21\\
\hline
\end{tabular}
}
\caption{\label{scorer}
F1 scores of SpanBERT\_base+c2f-coref on CRAFT-CR test set using different evaluation scripts.
}
\end{table*}

We run these models on the CRAFT-CR dataset of latest released version 4.0.1\footnote{\url{https://github.com/UCDenver-ccp/CRAFT/releases/tag/v4.0.1}}. CRAFT-CR consists of 97 full-text journal articles from PMC. As shown in Table \ref{statistics}, 60 documents are used for fine-tuning these models. 

These models are fine-tuned using learning rate of $1\times10^{-5}$ for PLMs parameters and $2\times10^{-4}$ for task parameters with Adam optimizer, a dropout of 0.3, and \verb|max_training_len| of 384 for SpanBERT\_base and 128 for other PLMs respectively. For SciBERT and PubMedBERT, we use the specific domain vocabulary, while general BERT vocabulary is used for other models.

For evaluation, we calculate F1 scores on six common metrics including $B^3$, BLANC, CEAFE, CEAFM, LEA and MUC using the official evaluation script\footnote{\url{https://github.com/UCDenver-ccp/craft-shared-tasks}} provided by the CRAFT shared task organizers, which is also used by previous models (\citealt{trieu2018investigating}, \citealt{trieu2019coreference}, \citealt{li2021knowledge}).

\subsubsection{Results}
The F1 scores of different PLMs combined with c2f-coref model on the CRAFT-CR test set are shown in Table \ref{experiment}. We can see that SpanBERT\_base achieved the best performance of 47.76\% F1 score, even without biomedical domain pre-training. This proves the powerful ability of SpanBERT on coreference resolution task.

In addition, biomedical PLMs outperform BERT\_base on CRAFT-CR, except SciBERT \citep{beltagy2019scibert} and UMLSBERT \citep{michalopoulos2021umlsbert}, which shows that biomedical domain knowledge can generally benefit coreference models for the biomedical domain. Moreover, Clinical KB-BERT \citep{hao2020enhancing}, which is initialized from BioBERT \citep{lee2020biobert}, achieved better performance than other biomedical PLMs, indicating that biomedical PLMs can be further enhanced by integrating external biomedical knowledge bases. However, SciBERT performs worse than BERT\_base on the CRAFT-CR dataset, although pre-trained on scientific texts and achieved better performance than BERT\_base on some other scientific NLP tasks such as NER, as reported in \citet{beltagy2019scibert}. One possible reason is that the pre-training corpora of SciBERT contain a number of computer science articles, which is unlikely to be beneficial for biomedical tasks.

Among these biomedical PLMs, SciBERT \citep{beltagy2019scibert} and PubMedBERT  \citep{gu2020domain} are pre-trained on domain-specific text from scratch, while others conduct continual pre-training based on the general domain. Although \citet{gu2020domain} shows that domain-specific pre-training from scratch outperforms continual pre-training from general-domain language models, the results of our experiment are the opposite.
Presumably the reason is that CRAFT-CR annotated all semantic type markables and covered a wider range of coreferences, so pre-training on the general domain is also beneficial.

\subsubsection{Results using different evaluation scripts}
Apart from the official evaluation script provided by CRAFT shared task organizers, we also used two other evaluation scripts, i.e., CoNLL scorer 9.0 and the CoVal script, to evaluate these models for comparing the differences between these evaluation scripts on the CRAFT-CR dataset. CoNLL scorer 9.0\footnote{\url{https://github.com/bill-baumgartner/reference-coreference-scorers}} is a modified version of the original reference coreference scorer \citep{pradhan2014scoring} used for CoNLL-2011/2012 shared tasks. It added an optional partial mention matching scheme and handling for discontinuous mentions, i.e., mentions composed of non-contiguous tokens. The CoVal script\footnote{\url{https://github.com/ns-moosavi/coval}} is a python coreference scorer for both CoNLL and ARRAU datasets \citep{uryupina2020annotating}.

The results of three different evaluation scripts for SpanBERT\_base+c2f-coref model on CRAFT-CR test set are shown in Table \ref{scorer}. The F1 scores of MUC, $B^3$ and CEAFE metrics as well as the averaged value are  provided. As shown in Table \ref{scorer}, a strange phenomenon is that the results of CoNLL scorer 9.0 are much lower than those of the other two evaluation scripts, especially on the $B^3$ and CEAFE metrics. The reason of that is not clear and needs further analysis. Whereas, the results of the CoVal script and CRAFT official evaluation script are close, although the scores of the latter are a little higher. 

In addition to the F1 scores, the time cost of the three evaluation scripts are quiet different. The CoNLL scorer 9.0 took about one and a half hours, while the CoVal script only needed about 30 seconds for evaluation.




\section{Conclusion}
\label{conclusion}
In this paper, we review and analyse the progress of biomedical coreference datasets, biomedical language representation models and coreference models for the biomedical domain. Biomedical coreference is an essential but challenging task. Some efforts have been made in this field, but there is still a much room for improvement. The experiments which we conducted indicate biomedical domain knowledge from either pre-training on biomedical texts or integrating biomedical knowledge bases can enhance coreference models for the biomedical domain.

\section*{Acknowledgements}
This research was supported in part by the China Scholarship Council, and the DALI project,
ERC Grant 695662.

\bibliography{anthology,custom}

\begin{thebibliography}{58}
\expandafter\ifx\csname natexlab\endcsname\relax\def\natexlab#1{#1}\fi

\bibitem[{Alsentzer et~al.(2019)Alsentzer, Murphy, Boag, Weng, Jindi, Naumann,
  and McDermott}]{alsentzer2019publicly}
Emily Alsentzer, John Murphy, William Boag, Wei-Hung Weng, Di~Jindi, Tristan
  Naumann, and Matthew McDermott. 2019.
\newblock Publicly available clinical bert embeddings.
\newblock In \emph{Proceedings of the 2nd Clinical Natural Language Processing
  Workshop}, pages 72--78.

\bibitem[{Batista-Navarro and Ananiadou(2011)}]{batista2011building}
Riza~Theresa Batista-Navarro and Sophia Ananiadou. 2011.
\newblock Building a coreference-annotated corpus from the domain of
  biochemistry.
\newblock In \emph{Proceedings of BioNLP 2011 Workshop}, pages 83--91.

\bibitem[{Beltagy et~al.(2019)Beltagy, Lo, and Cohan}]{beltagy2019scibert}
Iz~Beltagy, Kyle Lo, and Arman Cohan. 2019.
\newblock Scibert: A pretrained language model for scientific text.
\newblock In \emph{Proceedings of the 2019 Conference on Empirical Methods in
  Natural Language Processing and the 9th International Joint Conference on
  Natural Language Processing (EMNLP-IJCNLP)}, pages 3615--3620.

\bibitem[{Bodenreider(2004)}]{bodenreider2004unified}
Olivier Bodenreider. 2004.
\newblock The unified medical language system (umls): integrating biomedical
  terminology.
\newblock \emph{Nucleic acids research}, 32(suppl\_1):D267--D270.

\bibitem[{Castano et~al.(2002)Castano, Zhang, and
  Pustejovsky}]{castano2002anaphora}
Jos{\'e} Castano, Jason Zhang, and James Pustejovsky. 2002.
\newblock Anaphora resolution in biomedical literature.

\bibitem[{Choi et~al.(2014)Choi, Verspoor, and Zobel}]{choi2014evaluation}
Miji Choi, Karin Verspoor, and Justin Zobel. 2014.
\newblock Evaluation of coreference resolution for biomedical text.
\newblock In \emph{MedIR@ SIGIR}.

\bibitem[{Choi et~al.(2016)Choi, Zobel, and Verspoor}]{choi2016categorical}
Miji Choi, Justin Zobel, and Karin Verspoor. 2016.
\newblock A categorical analysis of coreference resolution errors in biomedical
  texts.
\newblock \emph{Journal of biomedical informatics}, 60:309--318.

\bibitem[{Clark et~al.(2019)Clark, Luong, Le, and Manning}]{clark2019electra}
Kevin Clark, Minh-Thang Luong, Quoc~V Le, and Christopher~D Manning. 2019.
\newblock Electra: Pre-training text encoders as discriminators rather than
  generators.
\newblock In \emph{International Conference on Learning Representations}.

\bibitem[{Cohen et~al.(2017)Cohen, Lanfranchi, Choi, Bada, Baumgartner,
  Panteleyeva, Verspoor, Palmer, and Hunter}]{cohen2017coreference}
K~Bretonnel Cohen, Arrick Lanfranchi, Miji Joo-young Choi, Michael Bada,
  William~A Baumgartner, Natalya Panteleyeva, Karin Verspoor, Martha Palmer,
  and Lawrence~E Hunter. 2017.
\newblock Coreference annotation and resolution in the colorado richly
  annotated full text (craft) corpus of biomedical journal articles.
\newblock \emph{BMC bioinformatics}, 18(1):1--14.

\bibitem[{Devlin et~al.(2019)Devlin, Chang, Lee, and
  Toutanova}]{devlin2019bert}
Jacob Devlin, Ming-Wei Chang, Kenton Lee, and Kristina Toutanova. 2019.
\newblock Bert: Pre-training of deep bidirectional transformers for language
  understanding.
\newblock In \emph{Proceedings of the 2019 Conference of the North American
  Chapter of the Association for Computational Linguistics: Human Language
  Technologies, Volume 1 (Long and Short Papers)}, pages 4171--4186.

\bibitem[{D'Souza and Ng(2012)}]{d2012anaphora}
Jennifer D'Souza and Vincent Ng. 2012.
\newblock Anaphora resolution in biomedical literature: a hybrid approach.
\newblock In \emph{Proceedings of the ACM Conference on Bioinformatics,
  Computational Biology and Biomedicine}, pages 113--122.

\bibitem[{Fei et~al.(2021)Fei, Ren, Zhang, Ji, and Liang}]{fei2021enriching}
Hao Fei, Yafeng Ren, Yue Zhang, Donghong Ji, and Xiaohui Liang. 2021.
\newblock Enriching contextualized language model from knowledge graph for
  biomedical information extraction.
\newblock \emph{Briefings in Bioinformatics}, 22(3):bbaa110.

\bibitem[{Gasperin et~al.(2007)Gasperin, Karamanis, and
  Seal}]{gasperin2007annotation}
Caroline Gasperin, Nikiforos Karamanis, and Ruth Seal. 2007.
\newblock Annotation of anaphoric relations in biomedical full-text articles
  using a domain-relevant scheme.
\newblock In \emph{Proceedings of DAARC}, volume 2007. Citeseer.

\bibitem[{Gasperin(2009)}]{gasperin2009statistical}
Caroline~V Gasperin. 2009.
\newblock Statistical anaphora resolution in biomedical texts.
\newblock Technical report, University of Cambridge, Computer Laboratory.

\bibitem[{Gu et~al.(2020)Gu, Tinn, Cheng, Lucas, Usuyama, Liu, Naumann, Gao,
  and Poon}]{gu2020domain}
Yu~Gu, Robert Tinn, Hao Cheng, Michael Lucas, Naoto Usuyama, Xiaodong Liu,
  Tristan Naumann, Jianfeng Gao, and Hoifung Poon. 2020.
\newblock Domain-specific language model pretraining for biomedical natural
  language processing.
\newblock \emph{arXiv preprint arXiv:2007.15779}.

\bibitem[{Hao et~al.(2020)Hao, Zhu, and Paschalidis}]{hao2020enhancing}
Boran Hao, Henghui Zhu, and Ioannis Paschalidis. 2020.
\newblock Enhancing clinical bert embedding using a biomedical knowledge base.
\newblock In \emph{Proceedings of the 28th international conference on
  computational linguistics}, pages 657--661.

\bibitem[{He et~al.(2020)He, Zhou, Xiao, Jiang, Liu, Yuan, and
  Xu}]{he2020integrating}
Bin He, Di~Zhou, Jinghui Xiao, Xin Jiang, Qun Liu, Nicholas~Jing Yuan, and Tong
  Xu. 2020.
\newblock Integrating graph contextualized knowledge into pre-trained language
  models.
\newblock In \emph{Proceedings of the 2020 Conference on Empirical Methods in
  Natural Language Processing: Findings}, pages 2281--2290.

\bibitem[{Hirschman(1997)}]{hirschman1997muc}
Lynette Hirschman. 1997.
\newblock Muc-7 coreference task definition, version 3.0.
\newblock \emph{Proceedings of MUC-7, 1997}.

\bibitem[{Hovy et~al.(2006)Hovy, Marcus, Palmer, Ramshaw, and
  Weischedel}]{hovy2006ontonotes}
Eduard Hovy, Mitch Marcus, Martha Palmer, Lance Ramshaw, and Ralph Weischedel.
  2006.
\newblock Ontonotes: the 90\% solution.
\newblock In \emph{Proceedings of the human language technology conference of
  the NAACL, Companion Volume: Short Papers}, pages 57--60.

\bibitem[{Joshi et~al.(2020)Joshi, Chen, Liu, Weld, Zettlemoyer, and
  Levy}]{joshi2020spanbert}
Mandar Joshi, Danqi Chen, Yinhan Liu, Daniel~S Weld, Luke Zettlemoyer, and Omer
  Levy. 2020.
\newblock Spanbert: Improving pre-training by representing and predicting
  spans.
\newblock \emph{Transactions of the Association for Computational Linguistics},
  8:64--77.

\bibitem[{Joshi et~al.(2019)Joshi, Levy, Zettlemoyer, and Weld}]{joshi2019bert}
Mandar Joshi, Omer Levy, Luke Zettlemoyer, and Daniel~S Weld. 2019.
\newblock Bert for coreference resolution: Baselines and analysis.
\newblock In \emph{Proceedings of the 2019 Conference on Empirical Methods in
  Natural Language Processing and the 9th International Joint Conference on
  Natural Language Processing (EMNLP-IJCNLP)}, pages 5803--5808.

\bibitem[{Kilicoglu and Demner-Fushman(2016)}]{kilicoglu2016bio}
Halil Kilicoglu and Dina Demner-Fushman. 2016.
\newblock Bio-scores: A smorgasbord architecture for coreference resolution in
  biomedical text.
\newblock \emph{PloS one}, 11(3):e0148538.

\bibitem[{Kim et~al.(2003)Kim, Ohta, Tateisi, and Tsujii}]{kim2003genia}
J-D Kim, Tomoko Ohta, Yuka Tateisi, and Jun’ichi Tsujii. 2003.
\newblock Genia corpus—a semantically annotated corpus for bio-textmining.
\newblock \emph{Bioinformatics}, 19(suppl\_1):i180--i182.

\bibitem[{Kim et~al.(2008)Kim, Ohta, and Tsujii}]{kim2008corpus}
Jin-Dong Kim, Tomoko Ohta, and Jun'ichi Tsujii. 2008.
\newblock Corpus annotation for mining biomedical events from literature.
\newblock \emph{BMC bioinformatics}, 9(1):1--25.

\bibitem[{Kim and Park(2004)}]{kim2004bioar}
Jung-Jae Kim and Jong~C Park. 2004.
\newblock Bioar: Anaphora resolution for relating protein names to proteome
  database entries.
\newblock In \emph{Proceedings of the Conference on Reference Resolution and
  Its Applications}, pages 79--86.

\bibitem[{Kim et~al.(2011)Kim, Riloff, and Gilbert}]{kim2011taming}
Youngjun Kim, Ellen Riloff, and Nathan Gilbert. 2011.
\newblock The taming of reconcile as a biomedical coreference resolver.
\newblock In \emph{Proceedings of BioNLP Shared Task 2011 Workshop}, pages
  89--93.

\bibitem[{Lee et~al.(2020)Lee, Yoon, Kim, Kim, Kim, So, and
  Kang}]{lee2020biobert}
Jinhyuk Lee, Wonjin Yoon, Sungdong Kim, Donghyeon Kim, Sunkyu Kim, Chan~Ho So,
  and Jaewoo Kang. 2020.
\newblock Biobert: a pre-trained biomedical language representation model for
  biomedical text mining.
\newblock \emph{Bioinformatics}, 36(4):1234--1240.

\bibitem[{Lee et~al.(2017)Lee, He, Lewis, and Zettlemoyer}]{lee2017end}
Kenton Lee, Luheng He, Mike Lewis, and Luke Zettlemoyer. 2017.
\newblock End-to-end neural coreference resolution.
\newblock In \emph{Proceedings of the 2017 Conference on Empirical Methods in
  Natural Language Processing}, pages 188--197.

\bibitem[{Lee et~al.(2018)Lee, He, and Zettlemoyer}]{lee2018higher}
Kenton Lee, Luheng He, and Luke Zettlemoyer. 2018.
\newblock Higher-order coreference resolution with coarse-to-fine inference.
\newblock In \emph{Proceedings of the 2018 Conference of the North American
  Chapter of the Association for Computational Linguistics: Human Language
  Technologies, Volume 2 (Short Papers)}, pages 687--692.

\bibitem[{Li et~al.(2018)Li, Rao, Zheng, and Zhang}]{li2018set}
Chen Li, Zhiqiang Rao, Qinghua Zheng, and Xiangrong Zhang. 2018.
\newblock A set of domain rules and a deep network for protein coreference
  resolution.
\newblock \emph{Database}, 2018.

\bibitem[{Li et~al.(2014)Li, Jin, Jiang, Zhang, and Huang}]{li2014coreference}
Lishuang Li, Liuke Jin, Zhenchao Jiang, Jing Zhang, and Degen Huang. 2014.
\newblock Coreference resolution in biomedical texts.
\newblock In \emph{2014 IEEE International Conference on Bioinformatics and
  Biomedicine (BIBM)}, pages 12--14. IEEE.

\bibitem[{Li et~al.(2021)Li, Ma, Zhou, Cheng, He, and Li}]{li2021knowledge}
Y~Li, X~Ma, X~Zhou, P~Cheng, K~He, and C~Li. 2021.
\newblock Knowledge enhanced lstm for coreference resolution on biomedical
  texts.
\newblock \emph{Bioinformatics (Oxford, England)}.

\bibitem[{Lin and Liang(2004)}]{lin2004pronominal}
Yu-Hsiang Lin and Tyne Liang. 2004.
\newblock Pronominal and sortal anaphora resolution for biomedical literature.
\newblock In \emph{Proceedings of the 16th Conference on Computational
  Linguistics and Speech Processing}, pages 101--109.

\bibitem[{Liu et~al.(2021)Liu, Shareghi, Meng, Basaldella, and
  Collier}]{liu2021self}
Fangyu Liu, Ehsan Shareghi, Zaiqiao Meng, Marco Basaldella, and Nigel Collier.
  2021.
\newblock Self-alignment pretraining for biomedical entity representations.
\newblock In \emph{Proceedings of the 2021 Conference of the North American
  Chapter of the Association for Computational Linguistics: Human Language
  Technologies}, pages 4228--4238.

\bibitem[{Michalopoulos et~al.(2021)Michalopoulos, Wang, Kaka, Chen, and
  Wong}]{michalopoulos2021umlsbert}
George Michalopoulos, Yuanxin Wang, Hussam Kaka, Helen Chen, and Alexander
  Wong. 2021.
\newblock Umlsbert: Clinical domain knowledge augmentation of contextual
  embeddings using the unified medical language system metathesaurus.
\newblock In \emph{Proceedings of the 2021 Conference of the North American
  Chapter of the Association for Computational Linguistics: Human Language
  Technologies}, pages 1744--1753.

\bibitem[{Miwa et~al.(2012)Miwa, Thompson, and Ananiadou}]{miwa2012boosting}
Makoto Miwa, Paul Thompson, and Sophia Ananiadou. 2012.
\newblock Boosting automatic event extraction from the literature using domain
  adaptation and coreference resolution.
\newblock \emph{Bioinformatics}, 28(13):1759--1765.

\bibitem[{Nguyen et~al.(2012)Nguyen, Kim, Miwa, Matsuzaki, and
  Tsujii}]{nguyen2012improving}
Ngan Nguyen, Jin-Dong Kim, Makoto Miwa, Takuya Matsuzaki, and Junichi Tsujii.
  2012.
\newblock Improving protein coreference resolution by simple semantic
  classification.
\newblock \emph{BMC bioinformatics}, 13(1):1--12.

\bibitem[{Nguyen et~al.(2011)Nguyen, Kim, and Tsujii}]{nguyen2011overview}
Ngan Nguyen, Jin-Dong Kim, and Jun'ichi Tsujii. 2011.
\newblock Overview of the protein coreference task in bionlp shared task 2011.
\newblock In \emph{Proceedings of the BioNLP Shared Task 2011 Workshop}, pages
  74--82. Citeseer.

\bibitem[{Peng et~al.(2019)Peng, Yan, and Lu}]{peng2019transfer}
Yifan Peng, Shankai Yan, and Zhiyong Lu. 2019.
\newblock Transfer learning in biomedical natural language processing: An
  evaluation of bert and elmo on ten benchmarking datasets.
\newblock In \emph{Proceedings of the 18th BioNLP Workshop and Shared Task},
  pages 58--65.

\bibitem[{Pradhan et~al.(2014)Pradhan, Luo, Recasens, Hovy, Ng, and
  Strube}]{pradhan2014scoring}
Sameer Pradhan, Xiaoqiang Luo, Marta Recasens, Eduard Hovy, Vincent Ng, and
  Michael Strube. 2014.
\newblock Scoring coreference partitions of predicted mentions: A reference
  implementation.
\newblock In \emph{Proceedings of the conference. Association for Computational
  Linguistics. Meeting}, volume 2014, page~30. NIH Public Access.

\bibitem[{Pustejovsky et~al.(2002)Pustejovsky, Castano, Sauri, Zhang, and
  Luo}]{pustejovsky2002medstract}
James Pustejovsky, Jos{\'e} Castano, Roser Sauri, Jason Zhang, and Wei Luo.
  2002.
\newblock Medstract: creating large-scale information servers from biomedical
  texts.
\newblock In \emph{Proceedings of the ACL-02 workshop on Natural language
  processing in the biomedical domain}, pages 85--92.

\bibitem[{raj Kanakarajan et~al.(2021)raj Kanakarajan, Kundumani, and
  Sankarasubbu}]{raj2021bioelectra}
Kamal raj Kanakarajan, Bhuvana Kundumani, and Malaikannan Sankarasubbu. 2021.
\newblock Bioelectra: Pretrained biomedical text encoder using discriminators.
\newblock In \emph{Proceedings of the 20th Workshop on Biomedical Language
  Processing}, pages 143--154.

\bibitem[{Segura-Bedmar et~al.(2010)Segura-Bedmar, Crespo,
  de~Pablo-S{\'a}nchez, and Mart{\'\i}nez}]{segura2010resolving}
Isabel Segura-Bedmar, Mario Crespo, C{\'e}sar de~Pablo-S{\'a}nchez, and Paloma
  Mart{\'\i}nez. 2010.
\newblock Resolving anaphoras for the extraction of drug-drug interactions in
  pharmacological documents.
\newblock In \emph{BMC bioinformatics}, volume~11, pages 1--9. BioMed Central.

\bibitem[{Shin et~al.(2020)Shin, Zhang, Bakhturina, Puri, Patwary, Shoeybi, and
  Mani}]{shin2020bio}
Hoo-Chang Shin, Yang Zhang, Evelina Bakhturina, Raul Puri, Mostofa Patwary,
  Mohammad Shoeybi, and Raghav Mani. 2020.
\newblock Bio-megatron: Larger biomedical domain language model.
\newblock In \emph{Proceedings of the 2020 Conference on Empirical Methods in
  Natural Language Processing (EMNLP)}, pages 4700--4706.

\bibitem[{Shoeybi et~al.(2019)Shoeybi, Patwary, Puri, LeGresley, Casper, and
  Catanzaro}]{shoeybi2019megatron}
Mohammad Shoeybi, Mostofa Patwary, Raul Puri, Patrick LeGresley, Jared Casper,
  and Bryan Catanzaro. 2019.
\newblock Megatron-lm: Training multi-billion parameter language models using
  model parallelism.
\newblock \emph{arXiv preprint arXiv:1909.08053}.

\bibitem[{Stoyanov et~al.(2010)Stoyanov, Cardie, Gilbert, Riloff, Buttler, and
  Hysom}]{stoyanov2010coreference}
Veselin Stoyanov, Claire Cardie, Nathan Gilbert, Ellen Riloff, David Buttler,
  and David Hysom. 2010.
\newblock Coreference resolution with reconcile.
\newblock In \emph{Proceedings of the ACL 2010 Conference Short Papers}, pages
  156--161.

\bibitem[{Su et~al.(2008)Su, Yang, Hong, Tateisi, and
  Tsujii}]{su2008coreference}
Jian Su, Xiaofeng Yang, Huaqing Hong, Yuka Tateisi, and Jun'ichi Tsujii. 2008.
\newblock Coreference resolution in biomedical texts: a machine learning
  approach.
\newblock In \emph{Dagstuhl Seminar Proceedings}. Schloss
  Dagstuhl-Leibniz-Zentrum f{\"u}r Informatik.

\bibitem[{Sung et~al.(2020)Sung, Jeon, Lee, and Kang}]{sung2020biomedical}
Mujeen Sung, Hwisang Jeon, Jinhyuk Lee, and Jaewoo Kang. 2020.
\newblock Biomedical entity representations with synonym marginalization.
\newblock In \emph{Proceedings of the 58th Annual Meeting of the Association
  for Computational Linguistics}, pages 3641--3650.

\bibitem[{Tateisi et~al.(2005)Tateisi, Yakushiji, Ohta, and
  Tsujii}]{tateisi2005syntax}
Yuka Tateisi, Akane Yakushiji, Tomoko Ohta, and Jun’ichi Tsujii. 2005.
\newblock Syntax annotation for the genia corpus.
\newblock In \emph{Companion Volume to the Proceedings of Conference including
  Posters/Demos and tutorial abstracts}.

\bibitem[{Torii and Vijay-Shanker(2005)}]{torii2005anaphora}
Manabu Torii and K~Vijay-Shanker. 2005.
\newblock Anaphora resolution of demonstrative noun phrases in medline
  abstracts.
\newblock In \emph{Proceedings of}, pages 332--339.

\bibitem[{Trieu et~al.(2019)Trieu, Nguyen, Nguyen, Miwa, Takamura, and
  Ananiadou}]{trieu2019coreference}
Hai-Long Trieu, Anh-Khoa~Duong Nguyen, Nhung Nguyen, Makoto Miwa, Hiroya
  Takamura, and Sophia Ananiadou. 2019.
\newblock Coreference resolution in full text articles with bert and
  syntax-based mention filtering.
\newblock In \emph{Proceedings of The 5th Workshop on BioNLP Open Shared
  Tasks}, pages 196--205.

\bibitem[{Trieu et~al.(2018)Trieu, Nguyen, Miwa, and
  Ananiadou}]{trieu2018investigating}
Hai~Long Trieu, Nhung~TH Nguyen, Makoto Miwa, and Sophia Ananiadou. 2018.
\newblock Investigating domain-specific information for neural coreference
  resolution on biomedical texts.
\newblock In \emph{Proceedings of the BioNLP 2018 workshop}, pages 183--188.

\bibitem[{Uryupina et~al.(2020)Uryupina, Artstein, Bristot, Cavicchio, Delogu,
  Rodriguez, and Poesio}]{uryupina2020annotating}
Olga Uryupina, Ron Artstein, Antonella Bristot, Federica Cavicchio, Francesca
  Delogu, Kepa~J Rodriguez, and Massimo Poesio. 2020.
\newblock Annotating a broad range of anaphoric phenomena, in a variety of
  genres: the arrau corpus.
\newblock \emph{Natural Language Engineering}, 26(1):95--128.

\bibitem[{Wu et~al.(2020)Wu, Wang, Yuan, Wu, and Li}]{wu2020corefqa}
Wei Wu, Fei Wang, Arianna Yuan, Fei Wu, and Jiwei Li. 2020.
\newblock Corefqa: Coreference resolution as query-based span prediction.
\newblock In \emph{Proceedings of the 58th Annual Meeting of the Association
  for Computational Linguistics}, pages 6953--6963.

\bibitem[{Yang et~al.(2004)Yang, Su, Zhou, and Tan}]{yang2004np}
Xiaofeng Yang, Jian Su, Guodong Zhou, and Chew~Lim Tan. 2004.
\newblock An np-cluster based approach to coreference resolution.
\newblock In \emph{COLING 2004: Proceedings of the 20th International
  Conference on Computational Linguistics}, pages 226--232.

\bibitem[{Ye et~al.(2020)Ye, Lin, Du, Liu, Li, Sun, and
  Liu}]{ye2020coreferential}
Deming Ye, Yankai Lin, Jiaju Du, Zhenghao Liu, Peng Li, Maosong Sun, and
  Zhiyuan Liu. 2020.
\newblock Coreferential reasoning learning for language representation.
\newblock In \emph{Proceedings of the 2020 Conference on Empirical Methods in
  Natural Language Processing (EMNLP)}, pages 7170--7186.

\bibitem[{Yuan et~al.(2021)Yuan, Liu, Tan, Huang, and
  Huang}]{yuan2021improving}
Zheng Yuan, Yijia Liu, Chuanqi Tan, Songfang Huang, and Fei Huang. 2021.
\newblock Improving biomedical pretrained language models with knowledge.
\newblock In \emph{Proceedings of the 20th Workshop on Biomedical Language
  Processing}, pages 180--190.

\bibitem[{Yuan et~al.(2020)Yuan, Zhao, and Yu}]{yuan2020coder}
Zheng Yuan, Zhengyun Zhao, and Sheng Yu. 2020.
\newblock Coder: Knowledge infused cross-lingual medical term embedding for
  term normalization.
\newblock \emph{arXiv preprint arXiv:2011.02947}.

\end{thebibliography}
\bibliographystyle{acl_natbib}

\end{document}